\documentclass[letterpaper, 10 pt, conference]{ieeeconf}  

\IEEEoverridecommandlockouts                              

\usepackage{xcolor}
\usepackage[linesnumbered,ruled,vlined]{algorithm2e}
\usepackage{graphicx}
\usepackage{multirow}
\usepackage[para,online,flushleft]{threeparttable}
\usepackage{ulem}
\usepackage{amsmath}
\usepackage{color}
\usepackage{amssymb}
\usepackage{booktabs}
\usepackage[colorlinks, linkcolor=red]{hyperref}
\usepackage{array}
\usepackage{float}
\usepackage{marvosym}
\SetKwInput{KwInput}{Input}                
\SetKwInput{KwOutput}{Output} 
\usepackage{textcomp}
\usepackage{pgfplots}
\usepackage{subcaption}

\pgfplotsset{width=10cm,compat=1.9}
\usetikzlibrary{
  pgfplots.groupplots,
  matrix
}
\usepackage{siunitx}

\newcolumntype{C}[1]{>{\centering\arraybackslash}p{#1}}

\title{\LARGE \bf
Embodiment-agnostic Action Planning via Object-Part Scene Flow}

\author{
Weiliang Tang, Jia-Hui Pan, Wei Zhan, Jianshu Zhou, Huaxiu Yao, \\ Yun-Hui Liu, Masayoshi Tomizuka, Mingyu Ding$^{\dag}$, and Chi-Wing Fu$^{\dag}$
\thanks{ \dag Corresponding authors. This work is supported by the InnoHK of the Government of Hong Kong via the Hong Kong Centre for Logistics Robotics, and the CUHK T Stone Robotics Institute. W. Tang, J.-H. Pan, and C.-W. Fu are with the Department of Computer Science and Engineering, the Chinese University of Hong Kong. 
W. Zhan, J. Zhou, M. Tomizuka, and M. Ding are with the Department of Mechanical Engineering, UC Berkeley. 
H. Yao is with the Department of Computer Science, UNC-Chapel Hill.
Y.-H. Liu is with the Department of Mechanical and Automation Engineering, The Chinese University of Hong Kong.}}

\begin{document}
\maketitle
\thispagestyle{empty}
\pagestyle{empty}
\section*{abstract}
Observing that the key for robotic action planning is to understand the target-object motion when its associated part is manipulated by the end effector,
we propose to generate the 3D object-part scene flow and extract its transformations to solve the action trajectories for 
diverse embodiments.
The advantage of our approach is that
it derives the robot action explicitly from object motion prediction, yielding a more robust policy by understanding the object motions.
Also, 
beyond policies trained on embodiment-centric data, our method is embodiment-agnostic, generalizable across diverse embodiments, and being able to learn from human demonstrations.
Our method comprises three components: an object-part predictor to locate the part for the end effector to manipulate, an RGBD video generator to predict future RGBD videos, and a trajectory planner to extract embodiment-agnostic transformation sequences and solve the trajectory for diverse embodiments.
Trained on videos even without trajectory data, our method still outperforms existing works significantly by \textbf{27.7\%} and \textbf{26.2\%} on the prevailing virtual environments MetaWorld and Franka-Kitchen, respectively. 
Furthermore, we conducted real-world experiments, showing that our policy, trained only with human demonstration, can be deployed to various embodiments.

\section{Introduction}

Traditionally, robot arm actions are hard-coded by human experts to execute pre-defined manipulation tasks in fixed environments. 
Such a control setting largely restricts the adaptation capabilities of robots to handle complex and dynamic environments.
To equip robots with autonomous and adaptable behaviors, various action planning approaches are proposed, e.g., imitation learning~\cite{levine2016end, finn2017deep, sun2018neural, brohan2023rt, padalkar2023open,chi2023diffusion, zhen20243d} that learn action plans based on manually-controlled demonstrations through supervised learning and reinforcement learning~\cite{yamada2021motion, yang2021hierarchical, nasiriany2022augmenting}, in which the action policies are learnt by trials and errors.

However, most existing approaches directly learn robotic actions for one or few specific end effectors, e,g, a two-finger gripper, a suction cup, or a dexterous hand, i.e., different embodiments.
%
%
On the one hand, they lack an understanding of the object motions to generate robotic actions, and so, these methods often overfit the training trajectories.
On the other hand, since different embodiments have different physical structures and interact with objects differently, these methods may overfit a specific embodiment and cannot be generalized to diverse robotic platforms in real-world scenarios.
%

Rethinking the robot manipulation, we observe that the key to successful manipulation is to understand the target object's motions. Hence, instead of directly generating robot actions, we plan robot actions based on the motions of the manipulated objects.
%
However, we cannot naively utilize the motions of the whole object, since the object may not be fully rigid. Hence, we exploit the motions of the target object part, i.e., considering a local object component that is approximately rigid and attaching it to the embodiment's end effector. 
Such part should share the same motion process as the embodiment's end effector, thus helping us to derive the embodiment's action.
This observation suggests that we should shift our focus from embodiment movements to understanding the target object part's movement.
Considering so enables us to unlock new possibilities in learning robust and embodiment-agnostic robotic policies.

\begin{figure}[t]
    \centering
    \includegraphics[width=0.99\linewidth]{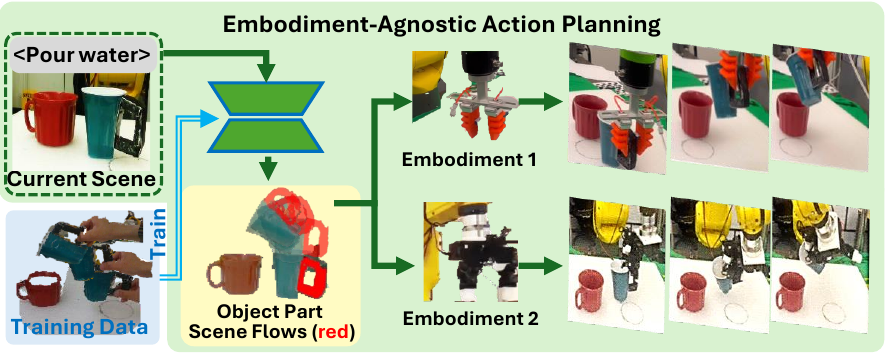}
    \vspace{-5pt}
    \caption{Illustration of our embodiment-agnostic action planning method. 
    Beyond existing approaches, our method learns to generate object-part scene flow (in red), independent of any specific embodiment, thereby enabling it to handle diverse embodiments and produce the execution trajectory.}
    \vspace{-13pt}
    \label{fig: teaser}
\end{figure}

Some recent works attempt embodiment-agnostic robot manipulation.
On the one hand,~\cite{garcia2020physics, mandikal2022dexvip, wang2024dexcap} pre-train models on mixed data from different embodiments to capture embodiment-irrelevant features. Yet, the embodiment types are limited for training and it is hard to generalize to new embodiments.
%
On the other hand,~\cite{mendonca2023structured, bahl2023affordances, goyal2022human, liu2022joint} extract embodiment-agnostic representations, however, they still require embodiment-related training procedures (behavior cloning, reinforcement learning, etc.) for deployment to achieve successful manipulation.
%
%
With advancements in flow analysis algorithms, some recent works~\cite{bharadhwaj2024track2act, yuan2024general, he2024large, xu2024flow} start to harness the optical flow to narrow the cross-embodiment gap in action planning.
However, optical flow depicts only 2D motion from the camera's view and struggles to model the target object's 3D motion.
Also, these methods utilize the optical flow of robots or the whole manipulated object, which is not sufficiently fine-grained to directly depict the accurate motion of the end effector.

Beyond the existing approaches, we propose a new embodiment-agnostic action-planning solution with two insights.
First, we learn to predict future motion specifically for the target object part for solving the robotic actions.
Second, we model the motion of the target object part by 3D scene flow, which is a 3D motion field of a set of points, each representing its instantaneous motion in 3D. 
As Fig.~\ref{fig: teaser} illustrates, we train the scene flow prediction using videos from diverse embodiments or even human hands.
Then, in the real-world deployment, we can feed the current observation to the trained model to obtain the scene flow and further extract the future motion of the object part.
So, for each specific embodiment, we can then apply the motion to obtain robust robotic actions for completing the task.

Procedure-wise, we decompose embodiment-agnostic action planning into three steps.
First, we recognize the target object part to be manipulated by the robot arm.
Second, we generate the future scene flow of the object part.
Lastly, we compute the motion of the object part from the predicted scene flow and explicitly solve for the end effector's trajectory relative to its initial pose.
Our approach achieves robust performance by trying to derive the end effector's trajectory from understanding object motion.
Moreover, our method is not limited to a single embodiment and can utilize human hand videos for training robotic policies.

To sum up, the contributions of this work include:
\begin{itemize}
    \item A new embodiment-agnostic action planning method by predicting the scene flow of the target object part.
    \item Our action-planning approach with an object-part predictor to locate the object part, a diffusion-based generator to forecast its scene flow, and an inverse action trajectory planner to generate robotic actions for any embodiment from the scene flow.
    \item We show our method's robustness by achieving state-of-the-art performance on two benchmarks, Meta-World and Franka-Kitchen, even without ground truths robot trajectory for training, significantly outperforming existing methods and baselines.
    \item Real-world experiments demonstrate that our method trained on human hand demonstrations can generalize to various embodiments effectively.
\end{itemize}


\section{Related Work}

\subsection{Learning Robot Manipulation from Videos}

Learning robot manipulation policies from videos have significantly expanded the capabilities of robotic systems. There are five branches of work.
The first branch~\cite{levine2016end, finn2017deep, sun2018neural, brohan2023rt, padalkar2023open,chi2023diffusion, zhen20243d} learns to map the visual observation to exact robotic actions directly. The second branch of work leverages action-value functions~\cite{liu2018imitation, sharma2019third,smith2019avid, edwards2019perceptual, schmeckpeper2020reinforcement} or extract reward functions from demonstrations~\cite{xiong2021learning, das2021model, sermanet2018time, scalise2019improving, pirk2019online} for reinforcement learning (RL). 
%
The third branch of work~\cite{garcia2020physics, mandikal2022dexvip} detects hand poses or key points from videos with humans and uses the poses or key points
to infer the robot arm's actions.
The fourth branch of work~\cite{nair2022r3m, sermanet2018time, he2024large, li2024ag2manip} focuses on extracting meaningful spatiotemporal representations from large-scale videos to achieve data-efficient RL learning. 
%
The last branch of work~\cite{du2024learning, wu2024ivideogpt, xu2024flow, bharadhwaj2024track2act} harnesses the diffusion model \cite{ho2020denoising} to generate manipulation videos and extracts partial robot actions from hallucinated image sequences.



\subsection{Embodiment-Agnostic Robot Manipulation}
Various approaches attempt to address embodiment-agnostic robot manipulation learning.
%
%
\cite{padalkar2023open, brohan2023rt, zhen20243d, kim2024openvla} pre-train models with a large amount of data with various embodiments to learn generalizable embeddings. 
Other methods extract embodiment-agnostic representations like affordance detection~\cite{mendonca2023structured, bahl2023affordances, goyal2022human, liu2022joint} and hand pose detection~\cite{garcia2020physics, mandikal2022dexvip, wang2024dexcap} to mitigate the embodiment gap. 
However, affordance and pose detection only predict the initial and goal scenes, and they still require embodiment-related training (RL, BC, etc.) to learn the intermediate motions.
%
With the recent advance of flow detection algorithms in computer vision, prevailing works~\cite{bharadhwaj2024track2act, yuan2024general, he2024large, xu2024flow} predict and harness the optical flow to infer robot actions in an embodiment-agnostic manner. 
However, these methods also require training on a specific embodiment for deployment, and they lack a 3D motion understanding of the specific manipulated object part.
%
%
Different from previous works, we generate robotic actions by learning the scene flow of the target object part, without needing any embodiment-related training data and offering a more comprehensive 3D motion understanding.


\begin{figure*}
    \centering
    \includegraphics[width=0.9\linewidth, height=0.3\linewidth]{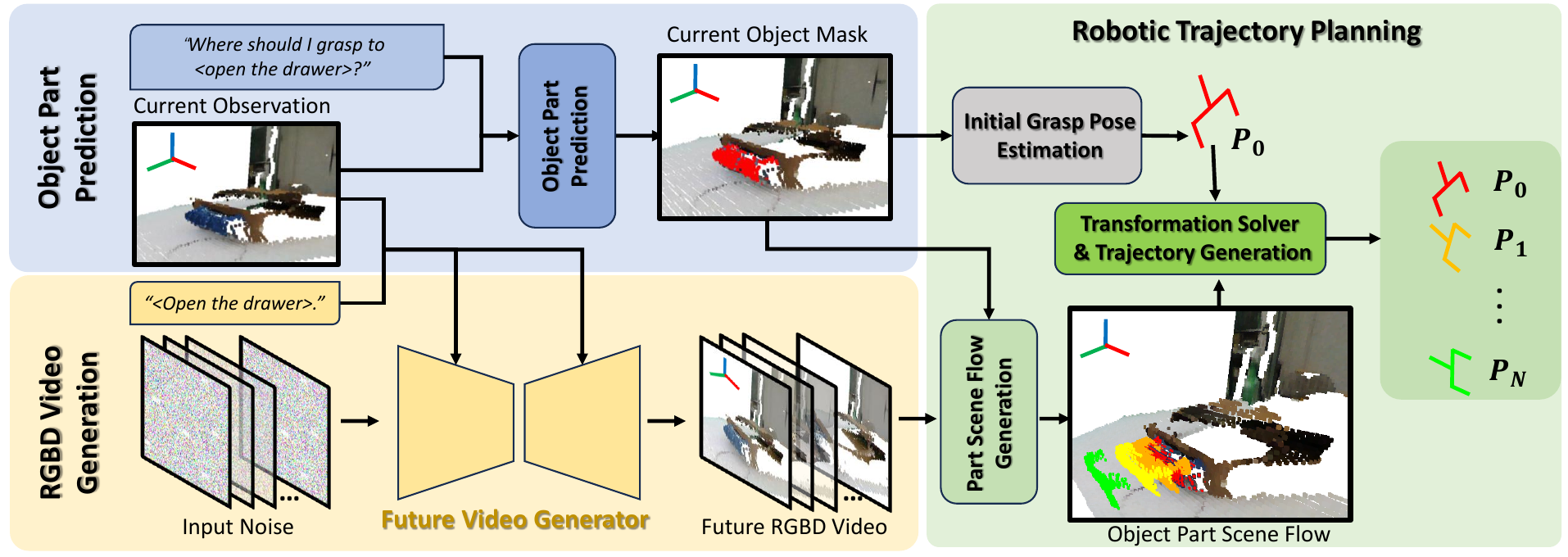}
    \caption{Framework overview. First, we identify the target object part and generate a future video to produce its hallucinated scene flow. Then, we predict the initial grasp pose and use a transformation solver on the scene flow for robotic trajectories.}
    \label{fig:framework}
\end{figure*}


\section{Method}
\subsection{Overview}
\label{subsec:problem_definition}
We are tasked to generate future robotic actions based on the current scene and a textual task description, ensuring that the process is embodiment-agnostic.
We achieve this by predicting future scene flow for the target object part and then extracting robotic actions inversely. 
%

An illustration of our framework is shown in Fig.~\ref{fig:framework}.
First, we predict the target object part via an object-part prediction module based on a vision-language model (VLM).
Then, we generate a future RGBD video using a text-guided diffusion model based on the current observation $o$ and the textural description of the task.
Next, we extract the scene flow of the target object part and solve the transformations of the object part between consecutive frames, 
thereby obtaining the end effector's motion.
%
Given an embodiment, we first predict its initial grasp pose $P_0$, and then use the object-part transformations to help generate the trajectory $P_{1:N}$ of the embodiment to complete the manipulation task.

\subsection{Object-part Prediction}
\label{subsec:object_part_prediction}


\begin{figure}
\centering
\begin{minipage}{0.11\textwidth}
  \centering
  \includegraphics[width=0.8\linewidth]{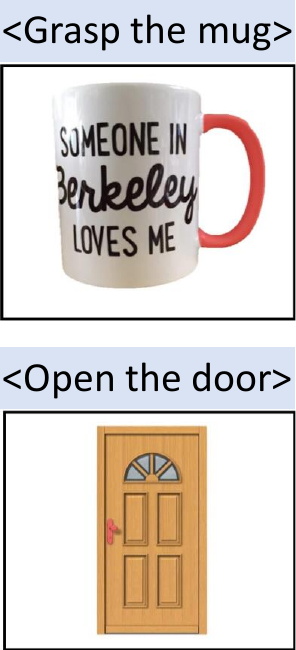}
  \caption{Object parts described by language are illustrated by mask (red).}
  \label{fig:part_seg}
\end{minipage} \hspace{+0.1pt}
\begin{minipage}{.36\textwidth}
  \centering
  \vspace{-2pt}
  \includegraphics[width=\linewidth]{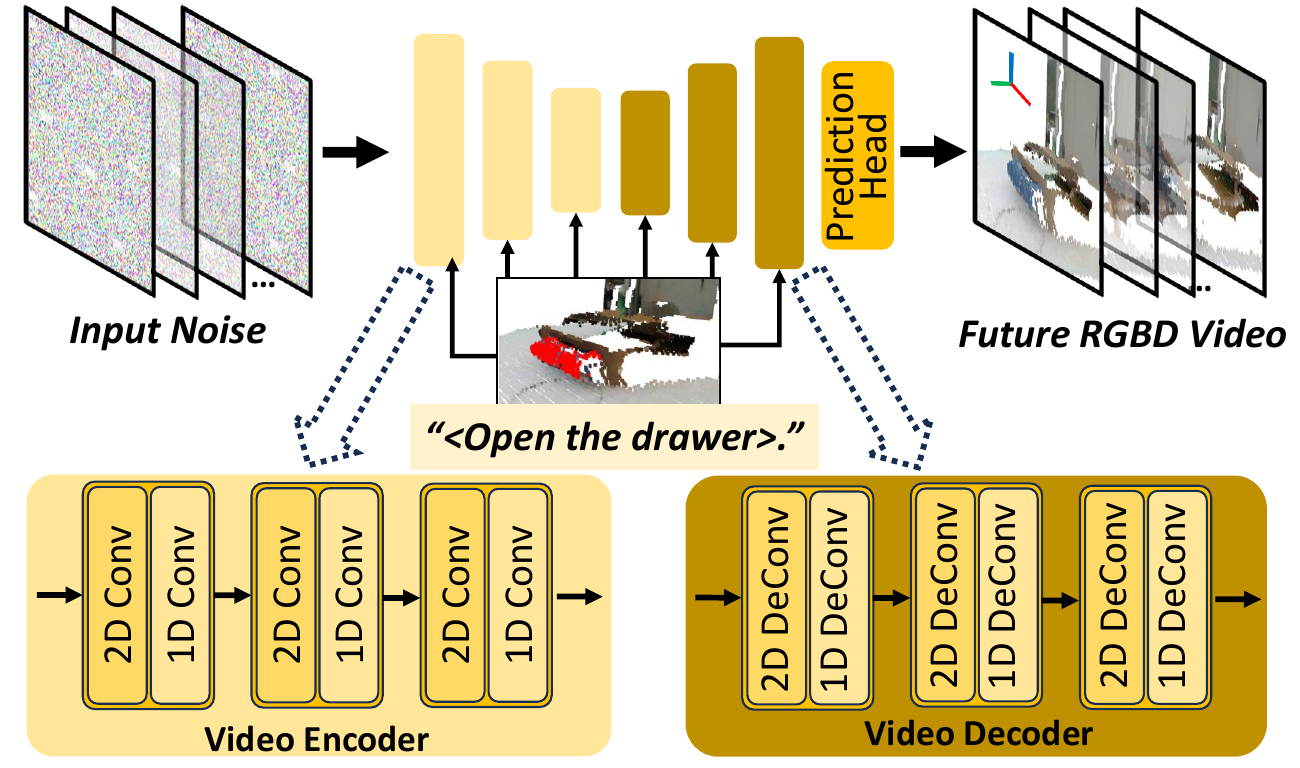}
  \caption{Illustration of our video generation network. The video generation network has a U-Net-like architecture and follows a denoising diffusion scheme to generate the RGBD sequence of the future frames.}
  \label{fig:network}
\end{minipage}
\vspace{-20pt}
\end{figure}

In robotic manipulation, a target object part is the component of the target object that the end effector directly interacts with.
It is typically rigid or small enough to be considered approximately rigid; however, it should have a medium size to ensure finding a stable grasp.
Predicting the target object part requires a basic understanding of the manipulation task, and which object part is usually grasped by a human or a robot to accomplish the task.
Thanks to the recent development of large vision-language models (VLMs)~\cite{radford2021learning,li2022blip,kim2021vilt,li2019visualbert}, some recent VLMs~\cite{lai2024lisa,ren2024grounded,li2022language} demonstrate a strong ability to understand language description and can accurately predict segmentation masks of object parts related to a task.

In this work, we adopt the LISA~\cite{lai2024lisa} model to locate the object part, which we denote as $F_{part}(\cdot)$.
Formally, we define the segmentation mask of the target object part as $S_{part} = F_{part}(O_{RGB}; \textbf{t})$, where $O_{RGB} \in \mathbb{R} ^{H\times W \times 3}$ is the RGB observation of the current frame, and $\textbf{t}$ is the textural prompt following the format \textit{``Where should I grasp if $<$task$>$? Please output segmentation masks."}.
The $<$task$>$ denotes the language description of the task we wish to perform, such as ``open the drawer", ``place the object on the shelf" and ``open the door".
The prediction $S_{part} \in \mathbb{R} ^{H\times W}$ is a 2D mask indicating the target object part, which is then used to extract its point clouds at the current frame to facilitate the future motion generation of the target object part.
See Fig.~\ref{fig:part_seg} which shows two examples of object parts predicted to ``grasp the mug" and ``open the door".

%
%


\subsection{Future Video Generation}
\label{subsec:RGBD_video_generation}
After obtaining the target object part, the next step is to predict its future motion.
Our goal is to manipulate the target object part to perform the desired action needed to complete the task. Because the target object part is rigidly attached to the end effector and moves with it, accurately predicting its motion is crucial for determining the robotic trajectory required for the task.
%
Thus, we develop a video generator to create future RGBD frames for task execution, which has a fundamental understanding of the current scene and the task we aim to perform.
Specifically, we leverage a text-conditioned RGBD video generator that takes the RGBD observation $O \in \mathbb{R}^{H\times W \times 4}$ of the current scene and the text description $<$task$>$ as input to generate $N$ future RGBD frames $I_{1:N} \in \mathbb{R} ^{H\times W \times 4 \times N}$, where $H$ and $W$ denote the height and width of each frame, respectively.
%

%
Inspired by~\cite{ko2023learning}, we develop our video generator with a U-Net~\cite{ronneberger2015u} architecture.
It consists of the same number of downsample blocks and upsample blocks. 
%
%
In each block, instead of using 3D convolution kernels, it alternatively applies a spatial convolution and a temporal convolution layer using a factorized spatial-temporal convolution kernel~\cite{sun2015human}.
%
%
%
To generate rich information to illustrate the dynamics of the scene, instead of directly generating the RGBD frames, we set the last upsample block as $C$, and further apply a prediction head to generate the predictions with the same channel size as the input. 
Fig.~\ref{fig:network} shows the network architecture of our video generator.

The video generator follows a denoising diffusion scheme to hallucinate future RGBD frames.
The input is initialized as Gaussian noise $\hat{I}^K_{1:N} \sim \mathcal N(0, \textbf{I})$ with a size of $H \times W \times 4 \times N$, and then the model progressively denoises the input for $K$ denoising step to yield the resulting RGBD video.
Instead of directly predicting the denoised image $\hat{I}_{1:N}^{k-1}$, we predict the noise added to $\hat{I}_{1:N}^{k-1}$ using the network, and we denote the network as $G_\epsilon(\cdot)$. The progressive generation is written as $\hat{I}^{k-1}_{1:N} = \hat{I}^k_{1:N} - G_\epsilon(\hat{I}^{k}_{1:N}; <task>, O, k)$, where $k \in [1, K]$ denotes the current denoising step. 
%
%
%
%
To train the model, we gather a set of videos showing any embodiment performing the task described in $<$task$>$.
This is because only the target object's part will be used to plan the robotic action, making it embodiment-agnostic.
We denote the ground-truth video frames as $I^{gt}_{1:N}$, and we leverage the MSE loss to train the video generator.
The training objective is written as:
\begin{equation}
    L_{MSE} = \|\epsilon - G_{\epsilon}(\sqrt{1 - \beta_k}\hat{I}^{k}_{1:N} + \sqrt{\beta_k}\epsilon; <task>, O, k)\|^2
\end{equation}
Where $\beta_t$ is the Gaussian noise schedule, $\epsilon$ is sampled from a multivariate standard Gaussian distribution, $I^{K}_{1:N}$ is the input noise at denoising step $k$, $O \in \mathbb{R} ^{H\times W \times 4}$ is the current observation of the scene, $G_{\epsilon}$ is the network to predict the added noise. 
After training $G_\epsilon(\cdot)$, we progressively yield the predicted RGBD video $I_{1:N} = \hat{I}^{0}_{1:N}$

\subsection{Trajectory Planning}
\label{subsec:traj_plan}
Given the hallucinate RGBD video $I_{1:N}$, and the segmentation mask of the object part $S_{part}$, we aim to generate a sequence of the end effector's poses $P_{1:N}$.
%
%
We accomplish this by first estimating the initial pose of the end effector $P_{0}$.
Next, we generate the scene flow for the target object part using $I_{1:N}$, and then extract a sequence of transformations $T_1, T_2, ..., T_n$ of the object-part motion.
The transformations are shared between the object part and the end effector because they are rigidly connected when the task is performed.
Consequently, we can apply the transformations on $P_{0}$ to yield the future poses of the end effector.

\vspace{+5pt}
\noindent\textbf{Initial Grasp Pose Estimation.}
We predict the end effector's initial pose $P_0$ 
using generalized grasp pose estimation methods.
%
Specifically, we leverage the open-sourced implementation of the grasp detection method~\cite{ten2017grasp}. 
Given the RGBD observation of the current frame $O$, and the segmentation mask $S_{part}$ of the target object part, we extract the pixels in $S_{part}$ and project them to the world coordinates to form the 3D point cloud $p_0 \in \mathbb{R} ^{M \times 3}$ at the current step, where $M$ denotes the number of points in $p_0$.
In addition, we generate the whole-scene point cloud $\hat{p}$.
Next, we generate grasp candidates through the grasp pose detection with object-part point cloud $p_0$.
Finally, we perform collision detection based on $\hat{p}$ to filter the pose candidates that collide with other objects to finally obtain the initial grasp pose $P_0$.


\vspace{+5pt}
\noindent\textbf{Scene Flow Generation.}
After obtaining the initial grasp pose of the end effector, the next step is to understand its future motion which is reflected by the target-object-part motion.
Therefore, we first generate the future scene flow of the target object part based on the  hallucinated RGBD videos $I_{1:N}$.
With the object-part segmentation mask $S_{part}$ and the object-part point cloud of the current scene already obtained, we can easily predict the object-part scene flow by tracking the motion of each point of the object part across consecutive frames in $I_{1:N}$.
%
%
%
In this work, we use the CoTracker~\cite{karaev2023cotracker}, which demonstrates superior performance of tracking any pixel in any videos.
More specifically, we feed the RGB video sequences into the CoTracker to track each pixel of the target object part throughout the future sequence.
By computing the motion vector of each pair of corresponding points between the pairs of point clouds $(p_{n-1}, p_{n})$ ($n \in [1, N]$), we form $s_{1:N} \in \mathbb{R}^{M \times 3 \times N}$, the scene flow of the object part in the $N$ future frames, where $M$ is the number of points in each object-part point cloud.
%

\vspace{+5pt}
\noindent\textbf{Trajectory Prediction.}
%
Based on our prior assumption that the target object part is rigidly attached to the end effector, the transformation of the end effector between consecutive frames is identical to that of the target object part.
Therefore, we can determine the end effector's trajectory by solving for the pose sequence of the object part $par$ in the future frames $1:N$.
Denoting the transformation matrix of $par$ from frame $n-1$ to frame $n$ as $T_n$ with $T_n \in \mathbb{R}^{4\times 4}$, we have $p'_{n} = T_n(p'_{n-1})$, where $p'_{n}, p'_{n-1} \in \mathbb{R}^{M \times 4}$ are the homogeneous coordinate of $p_{n}, p_{n-1} \in \mathbb{R}^{M \times 3}$, respectively, which is formed by $p'_{n} = concat(p_{n}, \textbf{1})$.
Similarly, we obtain a homogeneous version of the object-part scene flow $s'_{1:N}=concat(s_{1:N}, \textbf{0})$.
As a result, we can solve the relative transformations by
\begin{equation}
    \min_{T_n \in \textbf{SE}(3)} \|T_n p'_{n-1} - (p'_{n-1} + s'_{n-1})\|_2^2, n \in [1, N]
\end{equation}
where $\textbf{SE}(3)$ denotes the set of all rigid transformations in 3D space.
After the relative transformations, we resolve the trajectory of the end effector by $P_{n} = T_n(P_{n-1})$.
Note that when $n=1$, we use $P_0$ obtained through the initial grasp pose estimation.

\section{Experiment}


\begin{table*}
\centering
\small
\caption[]{\textbf{Results on the Meta-World Dataset.}
We report the average success rate of each method under tasks in the Meta-World dataset.
The best and the second-best success rates on each task are marked in bold and underlined, respectively.
The ``overall" denotes the average success rate across all tasks.
}
\vspace{-6pt}
\setlength{\tabcolsep}{3.1pt}
\resizebox{\linewidth}{!}{\begin{tabular}{lcccccccccccc}\toprule
& door-open & door-close & basketball & shelf-place & btn-press & btn-press-top & faucet-close & faucet-open & handle-press & hammer & assembly & \textbf{overall} \\
\midrule
BC-R3M & 1.3\% & 58.7\% & 0.0\% & 0.0\% & 36.0\% & 4.0\% & 18.7\% & 22.7\% & 28.0\% & 0.0\% & 0.0\% &  15.4\% \\
UniPi & 0.0\% & 36.0\% & 0.0\% & 0.0\% & 6.7\% & 0.0\% & 4.0\% & 9.3\% & 13.3\% & 4.0\% & 0.0\% & 6.1\% \\
Diffusion Policy & 45.3\% & 45.3\% & 8.0\% & 0.0\% & 40.0\% & 18.7\% & 22.7\% & \underline{58.7\%} & 21.3\% & 4.0\% &  1.3\% & 24.1\% \\
AVDC & \underline{72.0\%} & \underline{89.3\%} & \underline{37.3\%} & \underline{18.7\%} & \textbf{60.0}\% & \underline{24.0\%} & \underline{53.3\%} & 24.0\% & \textbf{81.3}\% & \underline{8.0\%} & \underline{6.7\%} & \underline{43.1\%} \\
\textbf{Ours} & \textbf{86.7}\% & \textbf{92.0}\% & \textbf{96.0}\% & \textbf{29.3}\% & \underline{50.7\%} & \textbf{96.0}\% & \textbf{100.0}\% & \textbf{64.0}\% & \underline{40.0\%} & \textbf{77.3}\% & \textbf{46.7}\% & \textbf{70.8}\% \\
\midrule
\textbf{Ours (GT Mask)} & \textbf{93.5}\% & \textbf{98.7}\% & \textbf{96.0}\% & \textbf{100.0}\% & \textbf{100.0}\% & \textbf{100.0}\% & \textbf{100.0}\% & \textbf{64.0}\% & \textbf{100.0}\% & \textbf{79.3}\% & \textbf{55.3}\% & \textbf{89.7}\% \\
\bottomrule
\end{tabular}}
\label{table:mw_main}
\end{table*}

\begin{table*}
\centering
\small
\caption[]{\textbf{Results on the Franka-Kitchen Dataset.}
We report the average success rate of each method under tasks in the Franka-Kitchen dataset.
The best and the second-best success rates on each task are marked in bold and underlined, respectively.
Considering that the results of the tuning-knob task are not accessible for ``VIP", ``Eureka" and ``Ag2Manip", we exclude the tuning-knob task when computer the overall performance across multiple tasks.
}
\vspace{-6pt}
\setlength{\tabcolsep}{8.8pt}
\scalebox{0.9}{\begin{tabular}{lccccccc}\toprule
& sliding-door & turning-light-on & opening-door & turning-knob & opening-microwave & \textbf{overall} (except turning-knob)\\
\midrule
BC-R3M & \textbf{82.5}\% & \underline{65.7}\% & 34.0\% & \underline{39.2\%} & \underline{34.8\%} & 54.3\%  \\
VIP & 0.0\% & 0.0\% & 0.0\% & -- & 0.0\% & 0.0\%  \\
Eureka & 0.0\% & 0.0\% & 0.0\% & -- & 0.0\% & 0.0\%  \\
Ag2Manip & 80.0\% & \textbf{70.0}\% & 70.0\% & -- & 10.0\% & \underline{57.5\%}  \\
\textbf{Ours} & \underline{82.0\%} & 60.0\% & \textbf{90.5}\% & \textbf{58.2}\% & \textbf{76.1}\% & \textbf{77.2}\%\\
\bottomrule
\end{tabular}}
\label{table:fk_main}
\vspace{-10pt}
\end{table*}

\subsection{Datasets and Implementation Details}
\vspace{+5pt}
\noindent\textbf{The Meta-World Dataset.}
Meta-World~\cite{yu2020meta} is a simulated benchmark with a Sawyer robot arm. 
This benchmark includes 11 distinct tasks, with RGBD video renderings obtained from 3 different camera perspectives for each task. 
%
%
We collect 5 demonstrations per task per camera perspective, resulting in 165 videos. Note that we discard the robot during rendering and our training data is robot-less.
Each policy is assessed across each task from the 3 camera perspectives, with each perspective undergoing 25 trials.

\vspace{+2pt}
\noindent\textbf{The Franka-Kitchen Dataset.}
Franka-Kitchen~\cite{gupta2019relay} is another simulated benchmark with the Franka Emika Panda robot arm. 
This benchmark includes 5 tasks, with RGBD video rendering from 2 different camera perspectives for each task. 
Following~\cite{nair2022r3m}, we collect demonstrations in 3 data sizes [25, 50, 100] per task per camera perspective and report the average performance. Note that our collected training data is also robot-less.
The policy is also assessed with 25 trials per task per camera perspective.

\vspace{+2pt}
\noindent\textbf{Evaluation Metrics.}
We evaluate the performance of varied methods on each task using the average success rate, in line with previous works~\cite{ko2023learning,du2024learning}.
A policy is successful if it achieves the goal state within the maximum number of environmental steps; otherwise, it is considered a failure.
The average success rate is calculated across all 25 trials within the same task. 

\vspace{+2pt}
\noindent\textbf{Implementation Details.} Due to the scarcity of training samples, we pre-train the model for Franka-Kitchen with 165 videos from Meta-World. Also, due to the lack of support for inverse kinematics, we adjust the robot base position in the environment so that the end effector can reach the object part in suitable poses. 
If a task fails to achieve within the maximum number of environment steps, we replan the policy starting from the failing point. The maximum replan times are set to 10.

\subsection{Results on the Meta-World Dataset}
We compare our method with four existing methods on the Meta-World dataset.
BC-R3M~\cite{nair2022r3m} is a BC method that generates robotic action based on the visual observations of the whole scene pre-trained with additional data~\cite{ko2023learning}.
%
Diffusion Policy~\cite{chi2023diffusion} infers future robotic actions using diffusion learning.
UniPi~\cite{du2024learning} predicts future image sequences and then infers robotic poses from them reversely.
The AVDC~\cite{ko2023learning} plans robotic actions by predicting future optical flow of the robot arms.
%

We report the results of our method and the compared methods in Tab.~\ref{table:mw_main}.
Considering that the ground-truth object-part mask in the virtual simulator is directly used in previous works~\cite{ko2023learning}, we additionally provide our result using the ground-truth object-part mask, marked as ``Ours (GT Mask)" in the table.
From the results we can see that our method can generate robust policy and greatly outperforms the compared methods by at least 27.7\% in terms of overall success rate averaged across all tasks even without using the ground-truth object-part mask.
Our method consistently achieves the best or second-best in each task, and the performance is even better when using the ground-truth object-part mask.

\subsection{Results on the Franka-Kitchen Dataset}
In the Franka-Kitchen dataset, we also compare our method with four state-of-the-art methods.
Eureka~\cite{ma2023eureka} uses RL and generates reward functions via large language models.
VIP~\cite{ma2022vip} uses value-implicit pre-training to generate rewards for RL training.
''Ag2Manip''~\cite{li2024ag2manip} trains a better representation for RL with point and trajectory information.
For fair comparisons, we use its version with a basic reward function.
Since the performance of some methods in the ``turning-knob" task is not accessible, we omit them and exclude them when computing the ``overall" performance across varied tasks.
We use the object masks extracted from the virtual environment. 

%

The results of our method and the compared methods are shown in Tab.~\ref{table:fk_main}.
From the results, we can see that our method also outperforms the compared methods on the Franka-Kitchen dataset.
It achieves the best or the second-best accuracy in each task and exceeds other methods by over 19.7\% in the overall performance.
The results further demonstrate the effectiveness of our method in embodiment-agnostic action planning for varied tasks.

\begin{table*}
\centering
\small

\setlength{\tabcolsep}{5.5pt}
\caption[]{\textbf{Ablation Results on the Franka-Kitchen Dataset.}
We report the average success rate of each ablating method under tasks in the Franka-Kitchen dataset.
The highest success rate for each task is marked in bold.}
\vspace{-6pt}
\scalebox{0.9}{\begin{tabular}{lccccccc}\toprule
& sliding-door & turning-light-on & opening-door & turning-knob & opening-microwave & \textbf{overall}\\
\midrule
\textbf{Ours (Full)} & \textbf{82.0}\% & \textbf{60.0}\% & \textbf{90.5}\% & \textbf{58.2}\% & \textbf{76.1}\% & \textbf{73.4}\%\\
w/o scene flow & 23.5\% & 6.0\% & 65.8\% & \textbf{58.2}\% & 50.0\% & 44.0\%\\
w/o scene flow \& object motion pred. & 20.0\% & 6.0\% & 43.5\% & 40.5\% & 31.6\% & 28.3\%\\
\bottomrule
\end{tabular}}
\label{table:ablate}
\vspace{-5pt}
\end{table*}

\subsection{Ablation Study}

\begin{figure}
    \centering
    \includegraphics[width=0.95\linewidth]{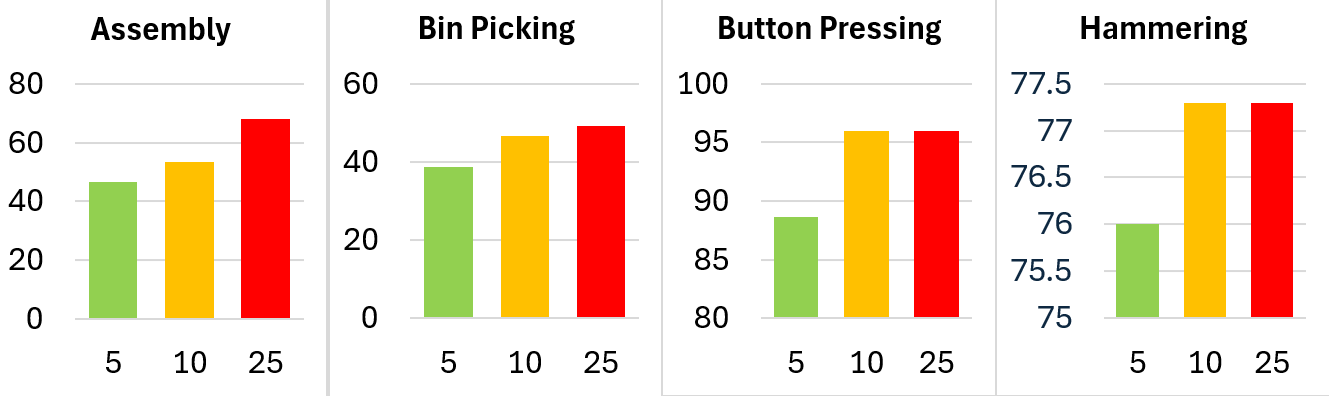}
    \vspace{-4pt}
    \caption{Success rates (\%, vertical axis) with respective to number of training demos (horizontal axis) in Meta-World.}
    \label{fig:number-demos}
\end{figure}

We perform an ablation study on the Franka-Kitchen dataset to evaluate the necessity of each designed module in our method.
%
The results are shown in Tab.~\ref{table:ablate}.
First, we use optical flow instead of scene flow (``w/o scene flow") to generate the end effector's action and predict the optical flow instead as in~\cite{ko2023learning}.
From the results, we can see that the performance drops due to the lack of 3D information to estimate the transformation of the object part.
%
%
%
Then, we further remove object motion prediction (``w/o scene flow \& object motion pred."), by using the whole RGBD scene as input and training a behavior cloning policy (BC) end-to-end to predict the trajectory of the end effector.
%
%
In addition, more training demos produce better object motion prediction, which leads to higher success rates as illustrated in Fig.~\ref{fig:number-demos}.

\subsection{Real World Demonstration}

\vspace{0pt}
\noindent \textbf{Experimental Settings.}
We use a FANUC LR mate 200id robot arm for real-world manipulation, and a RealSense D430 camera is fixed in front of the manipulation area. We collect RGBD image sequences of 30 human demonstration data for each task and use it to fine-tune the model pre-trained with the Meta-World dataset. Note that we don't fine-tune our model with any teleoperator data. As a comparison, we collect 30 ground-truth trajectories with a teleoperator and use them to train a behavior cloning model (BC). It uses ResNet-50~\cite{he2016deep} as the backbone to encode previous RGB input and generate action sequences. We use the LEAP hand~\cite{shaw2023leap} as the end effector. As our tasks only require grasp operation, we adjust the pose to mimic the two-fingered gripper. 
To test the cross-embodiment generalization, we use a different soft gripper~\cite{zhou2017soft, zhou2022bioinspired} and we illustrate that our method is generalizable for different embodiments.

\begin{figure}
    \centering
    \includegraphics[width=1.\linewidth]{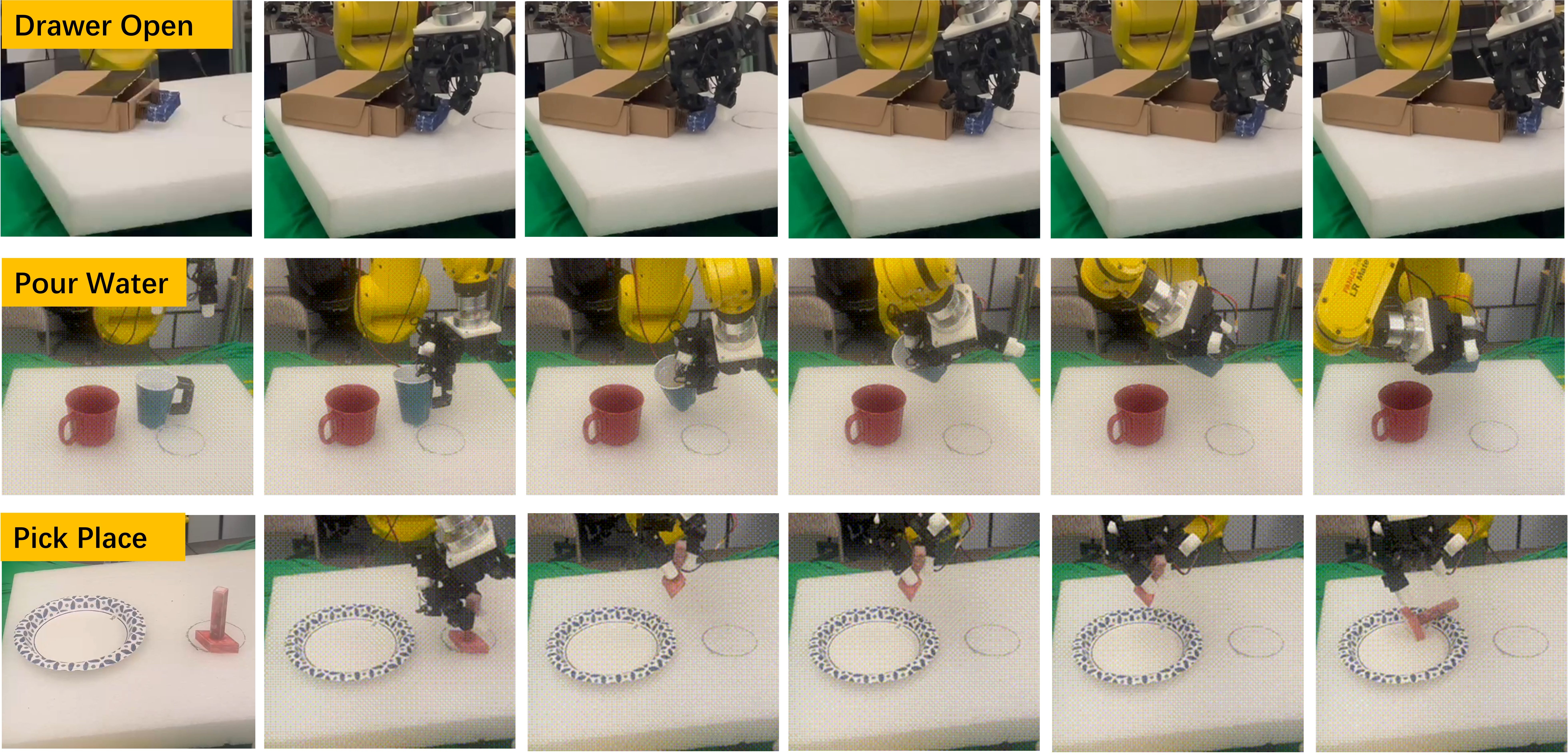}
    \caption{Visualization of the real-world manipulation results.}
    \vspace{-6pt}
    \label{fig:real-world}
\end{figure}

\vspace{+2pt}
\noindent \textbf{Experimental Results.}
Our performance is listed in Tab.~\ref{table:real-world-performance}. Examples of manipulation visualization are shown in Fig.~\ref{fig:real-world}. Due to the scarcity of training samples, the baseline method severely overfits trajectory and cannot generalize well, while our method can complete various tasks as expected.

\begin{figure}[t]
\begin{figure}[H]
    \centering
    \includegraphics[width=1.\linewidth]{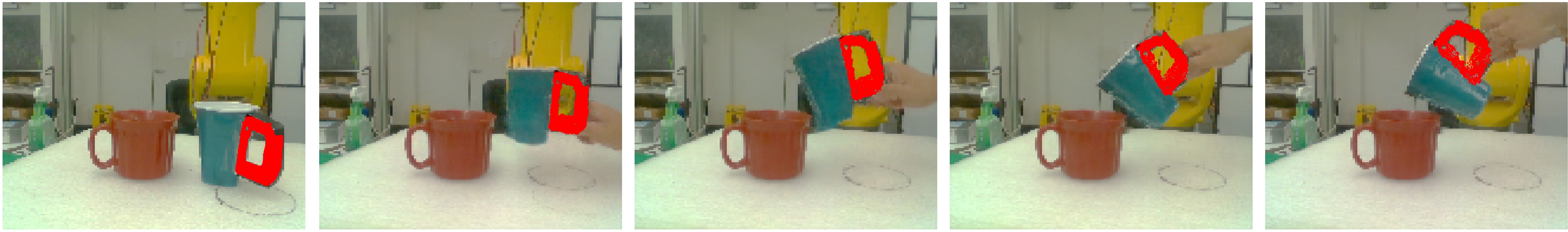}
    \caption{Although collected with the human hand and deployed on the robot, We can generate the correct image sequence for the object part for tracking. The part labeled in red is the tracked object part.}
    \vspace{-8pt}
    \label{fig:hand-embodiment}
\end{figure}

\begin{figure}[H]
    \centering
    \includegraphics[width=1.\linewidth]{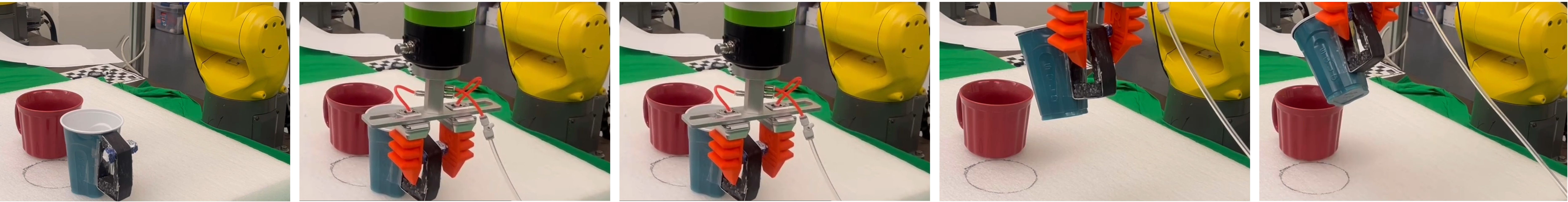}
    \caption{We generalize our method to different embodiments.}
    \vspace{-10pt}
    \label{fig:cross-embodiment}
\end{figure}
\end{figure}

\vspace{+2pt}
\noindent \textbf{Hand-Embodiment Generalization.}
\label{para: hand-emb-gap}
Since our method is object-hand-centric, we successfully avoid the hand-embodiment gap. As illustrated in Fig.~\ref{fig:hand-embodiment}, although our video is collected with human demonstration, when deployed with a robot gripper without a human hand, the object part is still successfully generated and we can track it correctly.

\begin{table}[t]
\centering
\setlength{\tabcolsep}{12.7pt}
\caption[]{\textbf{Success Rates on Real-World Tasks.}
We report the success rate of each method under 3 real-world tasks.
The highest success rate for each task is marked in bold.}
\scalebox{0.95}{\begin{tabular}{lccccc}\toprule
& open-drawer & pour-water & pick-place \\
\midrule
BC & 2/10  & 0/10 & 6/10 \\
\textbf{Ours} & 8/10& 5/10 & 9/10\\
\bottomrule
\end{tabular}}
\label{table:real-world-performance}
\vspace{-6pt}
\end{table}

\vspace{+2pt}
\noindent \textbf{Cross-Embodiment Generalization.}
\label{para: cross_embo}
We demonstrate that our method can achieve cross-embodiment performance. As illustrated in Fig.~\ref{fig:cross-embodiment}, we use a soft gripper~\cite{zhou2022bioinspired, zhou2017soft}, which is essentially different in shape, texture, and mechanism. Still, our method is irrelevant to embodiment and can generate correct trajectories and poses.

\section*{Conclusion}

This work introduces a novel embodiment-agnostic action planning framework that generates object-part scene flow and extracts transformations to solve action trajectories for any embodiment. 
The method achieves state-of-the-art performance and significantly outperforms existing methods and baselines on the Meta-World and Franka-Kitchen datasets.
Real-world experiments show that our approach, trained on human hand demonstrations, effectively generalizes across various embodiments.

Still, some limitations remain to be solved.
First, our method requires high-quality scene flow generation and object-part segmentation, and the performance can be further enhanced by using pre-trained models like Open-SORA~\cite{opensora} or refining the network design.
Secondly, while our method currently focuses on single manipulation tasks, future research could investigate complex long-horizon planning that combines multiple manipulation subtasks.


\clearpage
{
\small
\bibliographystyle{IEEEtran}
\bibliography{ref}

\begin{thebibliography}{10}
\providecommand{\url}[1]{#1}
\csname url@rmstyle\endcsname
\providecommand{\newblock}{\relax}
\providecommand{\bibinfo}[2]{#2}
\providecommand\BIBentrySTDinterwordspacing{\spaceskip=0pt\relax}
\providecommand\BIBentryALTinterwordstretchfactor{4}
\providecommand\BIBentryALTinterwordspacing{\spaceskip=\fontdimen2\font plus
\BIBentryALTinterwordstretchfactor\fontdimen3\font minus \fontdimen4\font\relax}
\providecommand\BIBforeignlanguage[2]{{%
\expandafter\ifx\csname l@#1\endcsname\relax
\typeout{** WARNING: IEEEtran.bst: No hyphenation pattern has been}%
\typeout{** loaded for the language `#1'. Using the pattern for}%
\typeout{** the default language instead.}%
\else
\language=\csname l@#1\endcsname
\fi
#2}}

\bibitem{levine2016end}
S.~Levine, C.~Finn, T.~Darrell, and P.~Abbeel, ``End-to-end training of deep visuomotor policies,'' \emph{Journal of Machine Learning Research}, vol.~17, no.~39, pp. 1--40, 2016.

\bibitem{finn2017deep}
C.~Finn and S.~Levine, ``Deep visual foresight for planning robot motion,'' in \emph{2017 IEEE International Conference on Robotics and Automation (ICRA)}.\hskip 1em plus 0.5em minus 0.4em\relax IEEE, 2017, pp. 2786--2793.

\bibitem{sun2018neural}
S.-H. Sun, H.~Noh, S.~Somasundaram, and J.~Lim, ``Neural program synthesis from diverse demonstration videos,'' in \emph{International Conference on Machine Learning}.\hskip 1em plus 0.5em minus 0.4em\relax PMLR, 2018, pp. 4790--4799.

\bibitem{brohan2023rt}
A.~Brohan, N.~Brown, J.~Carbajal, Y.~Chebotar, X.~Chen, K.~Choromanski, T.~Ding, D.~Driess, A.~Dubey, C.~Finn, \emph{et~al.}, ``Rt-2: Vision-language-action models transfer web knowledge to robotic control,'' \emph{arXiv preprint arXiv:2307.15818}, 2023.

\bibitem{padalkar2023open}
A.~Padalkar, A.~Pooley, A.~Jain, A.~Bewley, A.~Herzog, A.~Irpan, A.~Khazatsky, A.~Rai, A.~Singh, A.~Brohan, \emph{et~al.}, ``Open x-embodiment: Robotic learning datasets and rt-x models,'' \emph{arXiv preprint arXiv:2310.08864}, 2023.

\bibitem{chi2023diffusion}
C.~Chi, S.~Feng, Y.~Du, Z.~Xu, E.~Cousineau, B.~Burchfiel, and S.~Song, ``Diffusion policy: Visuomotor policy learning via action diffusion,'' \emph{arXiv preprint arXiv:2303.04137}, 2023.

\bibitem{zhen20243d}
H.~Zhen, X.~Qiu, P.~Chen, J.~Yang, X.~Yan, Y.~Du, Y.~Hong, and C.~Gan, ``3d-vla: A 3d vision-language-action generative world model,'' \emph{arXiv preprint arXiv:2403.09631}, 2024.

\bibitem{yamada2021motion}
J.~Yamada, Y.~Lee, G.~Salhotra, K.~Pertsch, M.~Pflueger, G.~Sukhatme, J.~Lim, and P.~Englert, ``Motion planner augmented reinforcement learning for robot manipulation in obstructed environments,'' in \emph{Conference on Robot Learning}.\hskip 1em plus 0.5em minus 0.4em\relax PMLR, 2021, pp. 589--603.

\bibitem{yang2021hierarchical}
X.~Yang, Z.~Ji, J.~Wu, Y.-K. Lai, C.~Wei, G.~Liu, and R.~Setchi, ``Hierarchical reinforcement learning with universal policies for multistep robotic manipulation,'' \emph{IEEE Transactions on Neural Networks and Learning Systems}, vol.~33, no.~9, pp. 4727--4741, 2021.

\bibitem{nasiriany2022augmenting}
S.~Nasiriany, H.~Liu, and Y.~Zhu, ``Augmenting reinforcement learning with behavior primitives for diverse manipulation tasks,'' in \emph{2022 International Conference on Robotics and Automation (ICRA)}.\hskip 1em plus 0.5em minus 0.4em\relax IEEE, 2022, pp. 7477--7484.

\bibitem{garcia2020physics}
G.~Garcia-Hernando, E.~Johns, and T.-K. Kim, ``Physics-based dexterous manipulations with estimated hand poses and residual reinforcement learning,'' in \emph{2020 IEEE/RSJ International Conference on Intelligent Robots and Systems (IROS)}.\hskip 1em plus 0.5em minus 0.4em\relax IEEE, 2020, pp. 9561--9568.

\bibitem{mandikal2022dexvip}
P.~Mandikal and K.~Grauman, ``Dexvip: Learning dexterous grasping with human hand pose priors from video,'' in \emph{Conference on Robot Learning}.\hskip 1em plus 0.5em minus 0.4em\relax PMLR, 2022, pp. 651--661.

\bibitem{wang2024dexcap}
C.~Wang, H.~Shi, W.~Wang, R.~Zhang, L.~Fei-Fei, and C.~K. Liu, ``Dexcap: Scalable and portable mocap data collection system for dexterous manipulation,'' \emph{arXiv preprint arXiv:2403.07788}, 2024.

\bibitem{mendonca2023structured}
R.~Mendonca, S.~Bahl, and D.~Pathak, ``Structured world models from human videos,'' \emph{arXiv preprint arXiv:2308.10901}, 2023.

\bibitem{bahl2023affordances}
S.~Bahl, R.~Mendonca, L.~Chen, U.~Jain, and D.~Pathak, ``Affordances from human videos as a versatile representation for robotics,'' in \emph{Proceedings of the IEEE/CVF Conference on Computer Vision and Pattern Recognition}, 2023, pp. 13\,778--13\,790.

\bibitem{goyal2022human}
M.~Goyal, S.~Modi, R.~Goyal, and S.~Gupta, ``Human hands as probes for interactive object understanding,'' in \emph{Proceedings of the IEEE/CVF Conference on Computer Vision and Pattern Recognition}, 2022, pp. 3293--3303.

\bibitem{liu2022joint}
S.~Liu, S.~Tripathi, S.~Majumdar, and X.~Wang, ``Joint hand motion and interaction hotspots prediction from egocentric videos,'' in \emph{Proceedings of the IEEE/CVF Conference on Computer Vision and Pattern Recognition}, 2022, pp. 3282--3292.

\bibitem{bharadhwaj2024track2act}
H.~Bharadhwaj, R.~Mottaghi, A.~Gupta, and S.~Tulsiani, ``Track2act: Predicting point tracks from internet videos enables diverse zero-shot robot manipulation,'' \emph{arXiv preprint arXiv:2405.01527}, 2024.

\bibitem{yuan2024general}
C.~Yuan, C.~Wen, T.~Zhang, and Y.~Gao, ``General flow as foundation affordance for scalable robot learning,'' \emph{arXiv preprint arXiv:2401.11439}, 2024.

\bibitem{he2024large}
H.~He, C.~Bai, L.~Pan, W.~Zhang, B.~Zhao, and X.~Li, ``Large-scale actionless video pre-training via discrete diffusion for efficient policy learning,'' \emph{arXiv preprint arXiv:2402.14407}, 2024.

\bibitem{xu2024flow}
M.~Xu, Z.~Xu, Y.~Xu, C.~Chi, G.~Wetzstein, M.~Veloso, and S.~Song, ``Flow as the cross-domain manipulation interface,'' \emph{arXiv preprint arXiv:2407.15208}, 2024.

\bibitem{liu2018imitation}
Y.~Liu, A.~Gupta, P.~Abbeel, and S.~Levine, ``Imitation from observation: Learning to imitate behaviors from raw video via context translation,'' in \emph{2018 IEEE international conference on robotics and automation (ICRA)}.\hskip 1em plus 0.5em minus 0.4em\relax IEEE, 2018, pp. 1118--1125.

\bibitem{sharma2019third}
P.~Sharma, D.~Pathak, and A.~Gupta, ``Third-person visual imitation learning via decoupled hierarchical controller,'' \emph{Advances in Neural Information Processing Systems}, vol.~32, 2019.

\bibitem{smith2019avid}
L.~Smith, N.~Dhawan, M.~Zhang, P.~Abbeel, and S.~Levine, ``Avid: Learning multi-stage tasks via pixel-level translation of human videos,'' \emph{arXiv preprint arXiv:1912.04443}, 2019.

\bibitem{edwards2019perceptual}
A.~D. Edwards and C.~L. Isbell, ``Perceptual values from observation,'' \emph{arXiv preprint arXiv:1905.07861}, 2019.

\bibitem{schmeckpeper2020reinforcement}
K.~Schmeckpeper, O.~Rybkin, K.~Daniilidis, S.~Levine, and C.~Finn, ``Reinforcement learning with videos: Combining offline observations with interaction,'' \emph{arXiv preprint arXiv:2011.06507}, 2020.

\bibitem{xiong2021learning}
H.~Xiong, Q.~Li, Y.-C. Chen, H.~Bharadhwaj, S.~Sinha, and A.~Garg, ``Learning by watching: Physical imitation of manipulation skills from human videos,'' in \emph{2021 IEEE/RSJ International Conference on Intelligent Robots and Systems (IROS)}.\hskip 1em plus 0.5em minus 0.4em\relax IEEE, 2021, pp. 7827--7834.

\bibitem{das2021model}
N.~Das, S.~Bechtle, T.~Davchev, D.~Jayaraman, A.~Rai, and F.~Meier, ``Model-based inverse reinforcement learning from visual demonstrations,'' in \emph{Conference on Robot Learning}.\hskip 1em plus 0.5em minus 0.4em\relax PMLR, 2021, pp. 1930--1942.

\bibitem{sermanet2018time}
P.~Sermanet, C.~Lynch, Y.~Chebotar, J.~Hsu, E.~Jang, S.~Schaal, S.~Levine, and G.~Brain, ``Time-contrastive networks: Self-supervised learning from video,'' in \emph{2018 IEEE international conference on robotics and automation (ICRA)}.\hskip 1em plus 0.5em minus 0.4em\relax IEEE, 2018, pp. 1134--1141.

\bibitem{scalise2019improving}
R.~Scalise, J.~Thomason, Y.~Bisk, and S.~Srinivasa, ``Improving robot success detection using static object data,'' in \emph{2019 IEEE/RSJ International Conference on Intelligent Robots and Systems (IROS)}.\hskip 1em plus 0.5em minus 0.4em\relax IEEE, 2019, pp. 4229--4235.

\bibitem{pirk2019online}
S.~Pirk, M.~Khansari, Y.~Bai, C.~Lynch, and P.~Sermanet, ``Online object representations with contrastive learning,'' \emph{arXiv preprint arXiv:1906.04312}, 2019.

\bibitem{nair2022r3m}
S.~Nair, A.~Rajeswaran, V.~Kumar, C.~Finn, and A.~Gupta, ``R3m: A universal visual representation for robot manipulation,'' \emph{arXiv preprint arXiv:2203.12601}, 2022.

\bibitem{li2024ag2manip}
P.~Li, T.~Liu, Y.~Li, M.~Han, H.~Geng, S.~Wang, Y.~Zhu, S.-C. Zhu, and S.~Huang, ``Ag2manip: Learning novel manipulation skills with agent-agnostic visual and action representations,'' \emph{arXiv preprint arXiv:2404.17521}, 2024.

\bibitem{du2024learning}
Y.~Du, S.~Yang, B.~Dai, H.~Dai, O.~Nachum, J.~Tenenbaum, D.~Schuurmans, and P.~Abbeel, ``Learning universal policies via text-guided video generation,'' \emph{Advances in Neural Information Processing Systems}, vol.~36, 2024.

\bibitem{wu2024ivideogpt}
J.~Wu, S.~Yin, N.~Feng, X.~He, D.~Li, J.~Hao, and M.~Long, ``ivideogpt: Interactive videogpts are scalable world models,'' \emph{arXiv preprint arXiv:2405.15223}, 2024.

\bibitem{ho2020denoising}
J.~Ho, A.~Jain, and P.~Abbeel, ``Denoising diffusion probabilistic models,'' \emph{Advances in neural information processing systems}, vol.~33, pp. 6840--6851, 2020.

\bibitem{kim2024openvla}
M.~J. Kim, K.~Pertsch, S.~Karamcheti, T.~Xiao, A.~Balakrishna, S.~Nair, R.~Rafailov, E.~Foster, G.~Lam, P.~Sanketi, \emph{et~al.}, ``Openvla: An open-source vision-language-action model,'' \emph{arXiv preprint arXiv:2406.09246}, 2024.

\bibitem{radford2021learning}
A.~Radford, J.~W. Kim, C.~Hallacy, A.~Ramesh, G.~Goh, S.~Agarwal, G.~Sastry, A.~Askell, P.~Mishkin, J.~Clark, \emph{et~al.}, ``Learning transferable visual models from natural language supervision,'' in \emph{International conference on machine learning}.\hskip 1em plus 0.5em minus 0.4em\relax PMLR, 2021, pp. 8748--8763.

\bibitem{li2022blip}
J.~Li, D.~Li, C.~Xiong, and S.~Hoi, ``Blip: Bootstrapping language-image pre-training for unified vision-language understanding and generation,'' in \emph{International conference on machine learning}.\hskip 1em plus 0.5em minus 0.4em\relax PMLR, 2022, pp. 12\,888--12\,900.

\bibitem{kim2021vilt}
W.~Kim, B.~Son, and I.~Kim, ``Vilt: Vision-and-language transformer without convolution or region supervision,'' in \emph{International conference on machine learning}.\hskip 1em plus 0.5em minus 0.4em\relax PMLR, 2021, pp. 5583--5594.

\bibitem{li2019visualbert}
L.~H. Li, M.~Yatskar, D.~Yin, C.-J. Hsieh, and K.-W. Chang, ``Visualbert: A simple and performant baseline for vision and language,'' \emph{arXiv preprint arXiv:1908.03557}, 2019.

\bibitem{lai2024lisa}
X.~Lai, Z.~Tian, Y.~Chen, Y.~Li, Y.~Yuan, S.~Liu, and J.~Jia, ``Lisa: Reasoning segmentation via large language model,'' in \emph{Proceedings of the IEEE/CVF Conference on Computer Vision and Pattern Recognition}, 2024, pp. 9579--9589.

\bibitem{ren2024grounded}
T.~Ren, S.~Liu, A.~Zeng, J.~Lin, K.~Li, H.~Cao, J.~Chen, X.~Huang, Y.~Chen, F.~Yan, \emph{et~al.}, ``Grounded sam: Assembling open-world models for diverse visual tasks,'' \emph{arXiv preprint arXiv:2401.14159}, 2024.

\bibitem{li2022language}
B.~Li, K.~Q. Weinberger, S.~Belongie, V.~Koltun, and R.~Ranftl, ``Language-driven semantic segmentation,'' \emph{arXiv preprint arXiv:2201.03546}, 2022.

\bibitem{ko2023learning}
P.-C. Ko, J.~Mao, Y.~Du, S.-H. Sun, and J.~B. Tenenbaum, ``Learning to act from actionless videos through dense correspondences,'' \emph{arXiv preprint arXiv:2310.08576}, 2023.

\bibitem{ronneberger2015u}
O.~Ronneberger, P.~Fischer, and T.~Brox, ``U-net: Convolutional networks for biomedical image segmentation,'' in \emph{Medical image computing and computer-assisted intervention--MICCAI 2015: 18th international conference, Munich, Germany, October 5-9, 2015, proceedings, part III 18}.\hskip 1em plus 0.5em minus 0.4em\relax Springer, 2015, pp. 234--241.

\bibitem{sun2015human}
L.~Sun, K.~Jia, D.-Y. Yeung, and B.~E. Shi, ``Human action recognition using factorized spatio-temporal convolutional networks,'' in \emph{Proceedings of the IEEE international conference on computer vision}, 2015, pp. 4597--4605.

\bibitem{ten2017grasp}
A.~Ten~Pas, M.~Gualtieri, K.~Saenko, and R.~Platt, ``Grasp pose detection in point clouds,'' \emph{The International Journal of Robotics Research}, vol.~36, no. 13-14, pp. 1455--1473, 2017.

\bibitem{karaev2023cotracker}
N.~Karaev, I.~Rocco, B.~Graham, N.~Neverova, A.~Vedaldi, and C.~Rupprecht, ``Cotracker: It is better to track together,'' \emph{arXiv preprint arXiv:2307.07635}, 2023.

\bibitem{yu2020meta}
T.~Yu, D.~Quillen, Z.~He, R.~Julian, K.~Hausman, C.~Finn, and S.~Levine, ``Meta-world: A benchmark and evaluation for multi-task and meta reinforcement learning,'' in \emph{Conference on robot learning}.\hskip 1em plus 0.5em minus 0.4em\relax PMLR, 2020, pp. 1094--1100.

\bibitem{gupta2019relay}
A.~Gupta, V.~Kumar, C.~Lynch, S.~Levine, and K.~Hausman, ``Relay policy learning: Solving long-horizon tasks via imitation and reinforcement learning,'' \emph{arXiv preprint arXiv:1910.11956}, 2019.

\bibitem{ma2023eureka}
Y.~J. Ma, W.~Liang, G.~Wang, D.-A. Huang, O.~Bastani, D.~Jayaraman, Y.~Zhu, L.~Fan, and A.~Anandkumar, ``Eureka: Human-level reward design via coding large language models,'' \emph{arXiv preprint arXiv:2310.12931}, 2023.

\bibitem{ma2022vip}
Y.~J. Ma, S.~Sodhani, D.~Jayaraman, O.~Bastani, V.~Kumar, and A.~Zhang, ``Vip: Towards universal visual reward and representation via value-implicit pre-training,'' \emph{arXiv preprint arXiv:2210.00030}, 2022.

\bibitem{he2016deep}
K.~He, X.~Zhang, S.~Ren, and J.~Sun, ``Deep residual learning for image recognition,'' in \emph{Proceedings of the IEEE conference on computer vision and pattern recognition}, 2016, pp. 770--778.

\bibitem{shaw2023leap}
K.~Shaw, A.~Agarwal, and D.~Pathak, ``Leap hand: Low-cost, efficient, and anthropomorphic hand for robot learning,'' \emph{arXiv preprint arXiv:2309.06440}, 2023.

\bibitem{zhou2017soft}
J.~Zhou, S.~Chen, and Z.~Wang, ``A soft-robotic gripper with enhanced object adaptation and grasping reliability,'' \emph{IEEE Robotics and automation letters}, vol.~2, no.~4, pp. 2287--2293, 2017.

\bibitem{zhou2022bioinspired}
J.~Zhou, H.~Cao, W.~Chen, S.~S. Cheng, and Y.-H. Liu, ``Bioinspired soft wrist based on multicable jamming with hybrid motion and stiffness control for dexterous manipulation,'' \emph{IEEE/ASME Transactions on Mechatronics}, vol.~28, no.~3, pp. 1256--1267, 2022.

\bibitem{opensora}
\BIBentryALTinterwordspacing
Z.~Zheng, X.~Peng, T.~Yang, C.~Shen, S.~Li, H.~Liu, Y.~Zhou, T.~Li, and Y.~You, ``Open-sora: Democratizing efficient video production for all,'' March 2024. [Online]. Available: \url{https://github.com/hpcaitech/Open-Sora}
\BIBentrySTDinterwordspacing

\end{thebibliography}
}

\end{document}